\documentclass[pmlr]{jmlr}

\usepackage{longtable}
\usepackage{multirow}

\usepackage{todonotes}

\usepackage{booktabs}

\usepackage{siunitx}
\usepackage{mathtools}

\usepackage{wrapfig}

\newcommand{\ie}{\textit{i.e.}}
\newcommand{\eg}{\textit{e.g.}}
\newcommand{\ddroit}{\mathrm{d}}
\newcommand{\xbold}{\mathbf{x}}

\theorembodyfont{\upshape}
\theoremheaderfont{\scshape}
\theorempostheader{:}
\theoremsep{\newline}

\jmlrvolume{277}
\jmlryear{2025}
\jmlrworkshop{Machine Learning Meets Differential Equations: From Theory to Applications}

\title[Learning non-Markovian Dynamical Systems with Signature-based Encoder]{Learning non-Markovian Dynamical Systems with Signature-based Encoders}

\makeatletter
\renewcommand{\Email}[1]{\hfill{\small\mdseries\upshape\scshape #1}}
\makeatother

\author{
    \Name{Eliott Pradeleix}        \Email{eliott.pradeleix@polytechnique.edu} \\
    \addr Ecole Polytechnique, 91120, Palaiseau France \\
    CNRS, Laboratoire Interdisciplinaire des Sciences du Numérique (LISN), Université Paris-Saclay, 91405, Orsay
    \AND
    \Name{Rémy Hosseinkhan-Boucher}
    \Email{remy.hosseinkhan@universite-paris-saclay.fr}
    \addr CNRS, Laboratoire Interdisciplinaire des Sciences du Numérique (LISN), Université Paris-Saclay, 91405, Orsay, France
    \AND
    \Name{Alena Shilova}
    \Email{alena.shilova@inria.fr} \\
    \addr Inria TAU, Laboratoire Interdisciplinaire des Sciences du Numérique (LISN), Université Paris-Saclay, 91405, Orsay, France
    \AND
    \Name{Onofrio Semeraro}
    \Email{onofrio.semeraro@cnrs.fr} \\
    \Name{Lionel Mathelin}
    \Email{lionel.mathelin@cnrs.fr} \\
    \addr CNRS, Laboratoire Interdisciplinaire des Sciences du Numérique (LISN), Université Paris-Saclay, 91405, Orsay}

\editors{Cecília Coelho, Bernd Zimmering, M.\ Fernanda P.\ Costa, Luís L.\ Ferrás, Oliver Niggemann}

\begin{document}

    \maketitle

    \begin{abstract}
        Neural ordinary differential equations offer an effective framework for modeling dynamical systems by learning a continuous-time vector field. However, they rely on the Markovian assumption—that future states depend only on the current state—which is often untrue in real-world scenarios where the dynamics may depend on the history of past states. 
        This limitation becomes especially evident in settings involving the continuous control of complex systems with delays and memory effects.
        To capture historical dependencies, existing approaches often rely on recurrent neural network (RNN)-based encoders, which are inherently discrete and struggle with continuous modeling. In addition, they may exhibit poor training behavior.
        In this work, we investigate the use of the signature transform as an encoder for learning non-Markovian dynamics in a continuous-time setting.
        The signature transform offers a continuous-time alternative with strong theoretical foundations and proven efficiency in summarizing multidimensional information in time.
        We integrate a signature-based encoding scheme into encoder-decoder dynamics models and demonstrate that it outperforms RNN-based alternatives in test performance on synthetic benchmarks. The code is available at : \url{https://github.com/eliottprdlx/Signature-Encoders-For-Dynamics-Learning.git}.
    \end{abstract}

    \section{Introduction}
    \label{sec:intro}

    Neural Ordinary Differential Equation (NODE), first introduced by \citet{chen2018neural}, allows one to model physical systems through an ODE of the form
    \begin{align*}
        \dot{\xbold}(t) = f_{\theta}(\xbold(t), t), \quad  \xbold(0) = \xbold_0
    \end{align*}
    with a learnable vector field 
    $f_{\theta}$,
    initial condition 
    $\xbold_0 \in \mathbb{R}^d$
    and state vector
    $\xbold(t) = (x_1(t), \, \dots, \allowbreak x_d(t)) \in \mathbb{R}^d$.
    Although effective in many situations, this framework fails when trying to model non-Markovian systems, \ie, systems whose future is influenced by their history, not just the present state, for instance 
    delayed differential equations \citep{kuang1993delay} of the form $\dot{\xbold}(t) = f(\xbold(t), \xbold(t-\tau), t)$.
    These non-Markovian differential equations naturally arise in biology \citep{gopalsamy1992stability, erneux2009applied, bressloff2013waves}, reinforcement learning and (stochastic) control theory \citep{holt2023control, hoglund2023neural}, especially under partial observability.
    
    To alleviate the representational limitations of NODEs, an augmented version was proposed by \citet{dupont2019augmented}, the so-called Augmented Neural ODE (ANODE).
    Letting $\mathbf{a}(t) \in \mathbb{R}^p$ denote a point in a vector space, the ODE problem is formulated as
    \begin{align*}
        \begin{bmatrix}
            \dot{\xbold}(t) \\\dot{\mathbf{a}}(t)
        \end{bmatrix}
        = f_\theta\left(\begin{bmatrix}
            \xbold(t) \\ \mathbf{a}(t)
        \end{bmatrix}, t\right), \quad \begin{bmatrix}
           \xbold(0) \\ \mathbf{a}(0)
        \end{bmatrix} =
        \begin{bmatrix}
            \mathbf{x_0} \\ \mathbf{0}
        \end{bmatrix}.
    \end{align*}
    While the additional dimensions introduced by ANODE may alleviate the bottleneck faced by NODE,
    \textit{i}) it comes at the cost of lifting to a higher dimensional space,
    \textit{ii}) the initial condition still depends solely on the current state,
    whereas explicit history-dependent information is required when modeling non-Markovian systems.
    In order to provide such information, several latent dynamics models have been proposed. These models aim to learn the vector field in a latent space, either by explicitly defining it in advance (\eg{}, using a Laplace-domain representation \citep{holt2022neural}) or by learning it implicitly from data.
    To do so, an encoder compresses history dependencies into a latent initial condition, and a decoder unrolls the trajectory to both reconstruct and extrapolate in the original space.
    In that spirit, \citet{rubanova2019latent} proposed Latent ODE to model the vector field in a latent space, and \citet{yildiz2019ode2vae} suggested to add higher-order terms.
    Nonetheless, these approaches still rely on the numerical solver of Neural ODE, which may face difficulties with certain classes of differential equations (\eg{}, stiff systems) \citep{holt2022neural}. 
    
    To model a broader class of differential equations, alternative models have been proposed \citep{holt2022neural, bilos2021neural}. The key idea is to bypass the numerical ODE solver by directly learning the latent vector \textit{flow}.
    These methods have proven to be very successful, although the choice of the encoder has been mostly overlooked.
    Indeed, the encoder is systematically chosen to be a recurrent neural network (RNN)-based model \citep{cho-etal-2014-learning, rubanova2019latent, debrouwer2019gruodebayes, bilos2021neural}.
    However, RNN are inadequate to deal with continuous time series because of their discrete nature.
    Moreover, they are computationally inefficient because they operate sequentially and are prone to vanishing gradient issues \citep{pascanu2013difficulty}.
    While ODE-GRU approaches \citep{rubanova2019latent, debrouwer2019gruodebayes} and GRU-Flow \citep{bilos2021neural} are better suited for continuous-time modeling, they still rely on the RNN architecture.
    It is also worth mentioning that studies on neural models constructed as delay differential equations have been conducted recently \citep{zhu2021neural, monsel2024neural, monsel2024time}.

    As an alternative, the signature transform introduced by \citet{chen1954iterated,chen1957integration,chen1958integration} has recently been used in several realms including rough path theory \citep{lyons1998differential,lyons2007differential,lyons2014rough}, finance, stochastic control \citep{morrill2021neural,sabatevidales2020solving,hoglund2023neural,perez2018derivatives}, and machine learning \citep{bonnier2019deep,fermanian2021learning,liao2021logsig,chevyrev2016primer,moreno2024rough}.
    The signature transform can be viewed as a collection of statistics that summarize the sequential structure of a trajectory, and it has proven to be a very effective tool to summarize the information of paths and dependencies across different dimensions, with high computational efficiency \citep{bonnier2019deep,reizenstein2020algorithm}.

    Moreover, numerous models leveraging the signature transform \citep{bonnier2019deep, kidger2020neuralcde, morrill2021neural, liao2021logsig, moreno2024rough} have shown increased performance in tasks involving continuous-time modeling in comparison to standard discrete RNNs and attention-based architectures.
    In addition, theoretical work \citep{fermanian2021learning} shows that there is a deep connection between RNN and signatures, the latter being in some sense a continuous-time counterpart of the former.

    In this work, we present a unified approach that integrates various models for learning dynamical systems within a general encoder-decoder framework.
    Then, we propose a signature-based encoder designed to learn from non-Markovian systems.
    Unlike RNN-based architectures that rely on discrete-time transitions, our encoding scheme is inherently suited for continuous-time modeling and may mitigate issues such as vanishing gradients, thereby facilitating more effective training.
    To support these claims, we empirically show that signature-based architectures exhibit better predicting performance and lead to faster training in comparison to their RNN-based counterparts.


    \section{Signature transform}
    \label{sec:background}

    The aim of this section is to provide the reader with a brief introduction to the signature transform and to give some intuition behind its use in the machine learning realm.
    \begin{definition}[$C^1$ Path]
        \label{def:c1-path}
        Let $a < b$ be two real numbers.  
        A \emph{$C^1$ path} is a continuously differentiable mapping $\xbold = (x_1,\dots,x_d) : t \in [a,b] \rightarrow \mathbb{R}^d$. We denote by $C^1([a,b]; \mathbb{R}^d)$ the set of such paths.
    \end{definition}
    We consider time series of the form $(\xbold({t_1}),\dots,\xbold({t_n})) \in \mathbb{R}^{d\times n}$, sampled from trajectories
    $(\xbold(t))_{a \leq t \leq b}$
    of typical dynamical system mentioned in the Introduction.
    Thus, we may see the vector $(\xbold({t_1}),\dots,\xbold({t_n})) \in \mathbb{R}^{d\times n}$ as being the time discretized version of a $C^1$ path $\xbold$. 
    \begin{definition}[Signature transform of $C^1$ paths]
        \label{def:c1-sig-transform}
        Let $\xbold \in C^1([a,b]; \mathbb{R}^d)$.
        The \textit{signature} of $\xbold$ is then defined as the collection of iterated integrals
        \begin{align*}
        \mathrm{Sig}(\xbold) &\coloneqq 
         \left(\left(
        \substack{
        \displaystyle \int \ldots \int \\
        \scriptstyle a < t_1 < \cdots < t_k < b
        }
        \prod_{j=1}^k \frac{\ddroit x_{i_j}}{\ddroit t}(t_j) \ddroit t_1 \cdots \ddroit t_k
        \right)_{1 \leq i_1, \ldots, i_k \leq d} \right)_{k \geq 0},
        \end{align*}
        where the $k = 0$ term is taken to be $1 \in \mathbb{R}$. The \textit{truncated signature} of depth $N$ of $\xbold$ is defined as
        \begin{align*}
        \mathrm{Sig}^N(\xbold) &\coloneqq 
         \left(\left(
        \substack{
        \displaystyle \int \ldots \int \\
        \scriptstyle a < t_1 < \cdots < t_k < b
        }
        \prod_{j=1}^k \frac{\ddroit x_{i_j}}{\ddroit t}(t_j) \ddroit t_1 \cdots \ddroit t_k
        \right)_{1 \leq i_1, \ldots, i_k \leq d} \right)_{0 \leq k \leq N}.
        \end{align*}
    \end{definition}
    For a complete understanding of the origin and the meaning of these iterated integrals, the reader is referred to~\cite{friz2010multidimensional}.

    Here, we introduce the signature transform only for $C^1$ paths for convenience, but it is actually defined on a broader class of functions (see Appendix~\ref{sec:appendix-sig-prop}), making it applicable to paths that are not strictly $C^1$, such as those arising in stiff differential equations.
    The signature transform provides a structured way to summarize sequential data by hierarchically extracting a collection of statistics from a path, thus offering a principled approach to deal with temporal data.
    Intuitively, one can think of the signature as an \textit{infinite dictionary} of path features, where simple statistics such as increments or areas under the curve appear at the first level, and higher-order terms capture more nuanced interactions.
    This layered representation makes this transformation particularly well-suited for machine learning applications, as it allows models to work with data in a compressed form yet rich in terms of information.
    We refer the reader to \citet{bonnier2019deep,fermanian2021learning,chevyrev2016primer} for further details on the use of signature methods in machine learning and the numerical computation of the signature transform.

    Briefly, the signature of a path captures its essential characteristics in a unique, robust to irregular-sampling, interpretable and efficient way, while being expressive enough so that any continuous function of the path can be arbitrarily approximated by a linear function of the signature, effectively acting as a universal nonlinearity.
    In addition, the magnitude of higher order terms behaves in a way that \textit{reasonable} depths are enough to essentially capture the information of the path.
    These claims are stated in a more formal way in Appendix \ref{sec:appendix-sig-prop}.

    \section{Encoder-decoder architectures for learning dynamical systems}
    \label{sec:encoder-decoder}

    Let $\mathcal{D}$ be a dataset consisting of time series
    $(\xbold({t_1}),\dots,\xbold({t_n})) \in \mathbb{R}^{d \times n}$
    sampled from trajectories
    $(\xbold(t))_{t \geq 0}$
    of some of the aforementioned dynamical systems.
    Let $u_\alpha : \mathbb{R}^{d \times n} \rightarrow \mathbb{R}^l$, $l \ll dn$,
    an encoder and
    $v_\beta : \mathbb{R}^l \times [0, +\infty[ \rightarrow \mathbb{R}^d$
    a decoder, respectively parameterized by $\alpha$, $\beta$ (\eg{}, neural networks).
    The goal of the encoder is to compress the history into a representation in the latent space, and the decoder can be interpreted as the composition of the flow in the latent space and the mapping back to the original space.

\begin{figure}
        \center
        \includegraphics[width=0.7\textwidth]{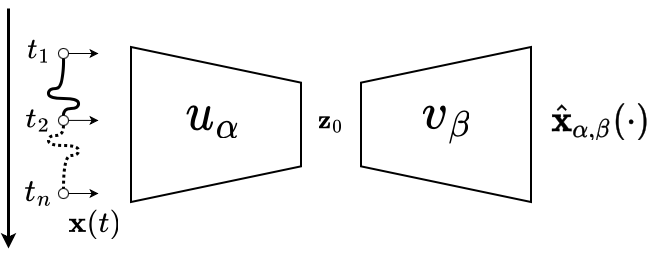}
        \caption{Encoder-decoder architecture for learning dynamical systems.}
        \label{fig:encoder-decoder}
        \vspace{-0.0cm}
    \end{figure}

        Within the context of learning dynamical systems, we define an encoder–decoder model as follows
    \[
        \hat{\xbold}_{\alpha,\beta}(t) \coloneqq v_\beta(u_\alpha(\textbf{x}({t_1}),...,\textbf{x}({t_n})), t).
    \]
   Such a model is trained on the dataset $\mathcal{D}$ using the mean squared error minimization objective
    \[
        \min_{\alpha, \beta} \sum_{(\xbold({t_1}),\dots,\xbold({t_n})) \in \mathcal{D}} \frac{1}{n}\sum_{i=1}^{n} || \xbold({t_i}) - \hat{\xbold}_{\alpha,\beta}(t_i) ||^2.
    \]
    Once trained, this architecture (Figure \ref{fig:encoder-decoder}) allows us to extrapolate and make continuous predictions on $\xbold({t > t_n})$.
    The most trivial example of encoder-decoder model is NODE, where the encoder is just a simple projection
    $u_{\alpha} : (\xbold({t_1}),\dots,\xbold({t_n})) \rightarrow \xbold({t_1})$\footnotemark[1],
    and the decoder is
    $v_{\beta} : (\mathbf{z}_0, t) \rightarrow \text{ODESolve}(f_\beta, \mathbf{z}_0, t)$
    with $f_\beta$ the vector field.
    For ANODE, the encoder becomes
    $u_{\alpha}: (\xbold({t_1}),\dots,\xbold({t_n}))
    \rightarrow
    \big[ \xbold(t_1)$\footnotemark[1]$ \, \, \, \mathbf{0} \big]$,
    and the decoder
    $v_{\beta}: (\mathbf{z}_0, t)
    \rightarrow 
    \pi(\text{ODESolve}(f_\beta, \mathbf{z}_0, t))$
    with $\pi$ being the projection on the first $d$ coordinates. \\
    \footnotetext[1]{or $\xbold({t_n)}$ for extrapolation (see Appendix \ref{sec:appendix-num-implementation} for further details).}

    \section{Methodology}
    \label{sec:method}

    \paragraph{Motivation} Architectures such as Neural ODEs \citep{chen2018neural}, Augmented Neural ODEs \citep{dupont2019augmented}, Latent ODEs/flows \citep{rubanova2019latent,bilos2021neural}, ODE2VAE \citep{yildiz2019ode2vae} and Neural Laplace \citep{holt2022neural} all fall under the general framework discussed in \sectionref{sec:encoder-decoder}, with different designs for $u_\alpha$ and $v_\beta$.
    In all cases, the encoder is either a simple coordinate projection or a RNN-based model.
    In the following, we propose a signature-based model for $u_\alpha$.

    \paragraph{Signature-based encoder} We now present a signature-based encoder, compatible with the aforementioned dynamics learning models, and largely inspired from the literature \citep{bonnier2019deep,liao2021logsig,moreno2024rough}.
    The intuition behind such a choice of encoding scheme is multifaceted. Indeed, the goal of the encoder is to perform feature selection on paths. Signature-based models : \textit{i}) have proven to be excellent path descriptors and exhibit universality
    \citep{fermanian2021learning,bonnier2019deep}, \textit{ii}) are particularly suited for continuous-time modeling, especially to jointly account for multidimensional correlations \citep{bonnier2019deep,liao2021logsig,moreno2024rough}, \textit{iii}) are more computationally efficient \citep{bonnier2019deep, reizenstein2020algorithm} and interpretable than standard RNNs.  

    Let $(\xbold({t_1}),\dots,\xbold({t_n})) \in \mathbb{R}^{d \times n}$ be a stream of data.
    Let $\phi_{\theta} : \mathbb{R}^{d \times m} \rightarrow \mathbb{R}^e$, $m < n$, a feed-forward neural network.
    As proposed by \citet{bonnier2019deep}, we first apply the mapping
    \[
        \Phi_{\theta}(\xbold({t_1}),\dots,\xbold({t_n})) \coloneqq (\phi_{\theta}(\xbold({t_1}),\dots,\xbold({t_m})),\dots,\dots,\phi_{\theta}(\xbold({t_{n-m+1}}),\dots,\xbold({t_n}))),
    \]
    to our stream of data, akin to 1D convolution.
    The rationale is that, since the signature must be truncated and an appropriate choice of depth is \textit{a priori} unknown, this approach may alleviate issues that arise when path information depends nontrivially on higher-order signature terms \citep{bonnier2019deep}.
    A vector $\Phi_{\theta}(\xbold({t_1}),\dots,\xbold({t_n})) \in \mathbb{R}^{e \times (n-m+1)}$ is obtained.
    Then, we apply the
    $N^{th}$-truncated signature to get a vector 
    $(\mathrm{Sig}^N \circ \Phi_{\theta})
    (\xbold({t_1}),\dots,\xbold({t_n}))
    \in \mathbb{R}^{q}$
    with $q = \frac{e^{N+1}-1}{e-1}$.
    Finally, we need a projection mapping $g_\xi : \mathbb{R}^{q} \rightarrow \mathbb{R}^l$, with $l$ the dimension of the latent space.
    In practice, we choose it to be a feed-forward neural network. \textit{In fine}, the encoder is defined as
    \[
        u_\alpha(\xbold({t_1}),...,\xbold({t_n})) \coloneqq (g_\xi \circ \mathrm{Sig}^N\circ\Phi_{\theta})(\xbold({t_1}),\dots,\xbold({t_n}))
    \]
    with $\alpha = (\theta, \xi)$.


    \section{Numerical experiments}
    \label{sec:numexp}

    A key aim of this study is to provide evidence of the performance improvements achieved through the introduction of the signature into the encoder architecture.
    We evaluate our signature-based encoder on a diverse set of dynamical systems, encompassing a variety of dimensions, delays, stiffness, and chaotic behaviors, with applications in biology, healthcare, and engineering. 
    For each system, our encoder is tested within two distinct model architectures-Neural Laplace~\citep{holt2022neural} and Neural Flow ResNet~\citep{bilos2021neural} (see Appendix~\ref{sec:appendix-models-description} for detailed model descriptions) and compared with its RNN-based counterpart used in the original paper. In what follows, Sig Neural Laplace (resp. Sig Neural Flow ResNet) denotes the model where the original encoder described in Appendix~\ref{sec:appendix-models-description} is replaced with the signature-based encoder from Section~\ref{sec:method}.
    
    To test the performance of the models, we use the exact same method as \citet{holt2022neural}. We evaluate model performance in an extrapolation setting by splitting each sampled trajectory into two parts $[0, T/2]$ and $[T/2, T]$.
    The first half is used for encoding, and the second half for prediction. 
    For each dynamical system, we sample 1,000 trajectories, each initialized from a distinct initial condition and consisting of 200 time points. To ensure a fair comparison focused on the choice of encoder, we keep the decoder architecture constant in terms of parameter count across all configurations, while ensuring that the signature-based encoder consistently uses fewer parameters than its RNN-based counterpart
    (see Table \ref{tab:param_counts}).
    For further details, see Appendix \ref{sec:appendix-num-implementation} and \ref{sec:appendix-sampling-de}. 

    \subsection{Description of test datasets}
    \label{subsec:desc-datasets}

    \paragraph{Delayed Lotka-Volterra} We consider a delayed variant of the classical Lotka--Volterra system \citep{gopalsamy1992stability, kuang1993delay} which incorporates a fixed delay capturing realistic biological lags.
    The model writes
    \begin{align*}
        \dot{x}_1(t) &= x_1(t)\left(1 - x_2(t - \tau)\right), \\
        \dot{x}_2(t) &= \frac{1}{2} x_2(t)\left(1 - x_1(t - \tau)\right)
    \end{align*}
    with $\tau > 0$ the delay.

    \paragraph{Spiral DDE} Spiral delay differential equations \citep{zhu2021neural} arise in biological and healthcare systems, capturing nonlinear delayed feedback where the dynamics with saturation effect introduced by the hyperbolic tangent.
    The model writes
    \begin{align*}
        \dot{\xbold}(t) = A \tanh(\xbold(t) + \xbold(t - \tau))
    \end{align*}
    with $\xbold(t) \in \mathbb{R}^2$, $\tau > 0$ the delay and $A$ is a $2 \times 2$ matrix with real coefficients.

    \paragraph{Delayed Fitzhugh-Nagumo}
    The delayed Fitzhugh-Nagumo dynamical system \citep{erneux2009applied, bressloff2013waves} models the interaction between a fast activator $x_1$, and a slow recovery variable $x_2$.
    The system writes
    \begin{align*}
        \dot{x}_1(t) &= x_1(t) - \frac{x_1(t)^3}{3} - x_2(t - \tau) + I, \\
        \dot{x}_2(t) &= \varepsilon \left(x_1(t) + a - b x_2(t) \right)
    \end{align*}
    where $a, b, I > 0$ and $\epsilon > 0$ are system parameters, and $\tau > 0$ the delay.
    The system is known to exhibit stiffness when $\epsilon \ll 1$.

    \paragraph{Delayed Rössler}
    The delayed Rössler dynamical system extends the classical Rössler model \citep{rossler1976equation} of chaotic dynamics by incorporating time-delayed feedback, which enables the investigation of broader temporal behaviors.
    The model writes
    \begin{align*}
        \dot{x}_1(t) &= -x_2(t) - x_3(t), \\
        \dot{x}_2(t) &= x_1(t) + a x_2(t), \\
        \dot{x}_3(t) &= b +x_3(t) \left[ x_1(t - \tau) - c \right],
    \end{align*}
    where $a, b, c > 0$ are system parameters and $\tau > 0$ the delay.
    The classical system is known to be chaotic for some well-chosen parameters.

    \subsection{Results}
    \label{subsec:results}

    \begin{table}[!ht]
    \floatconts
    {tab:rmse-full-annotated}
    {\caption{Test RMSE (mean $\pm$ std) averaged across 5 runs (random seed initialization). Best results per group bolded. NODE and ANODE are used as baselines.}}
    {%
        {\small
        \renewcommand{\arraystretch}{1.2} 
        \setlength{\tabcolsep}{3pt} 
        \begin{tabular}{lcccc}
            \toprule
            Method &
            \shortstack{Delayed \\Lotka-Volterra} &
            \shortstack{Spiral \\DDE} &
            \shortstack{Delayed \\Fitzhugh-Nagumo} &
            \shortstack{Delayed \\Rössler}  \\
            \midrule
            ANODE   & $.4472 \pm .0642$ & $.0435 \pm .0070$ & $.1001 \pm .0531$ & $1.3104 \pm .2578$ \\
            NODE & $.6061 \pm .2521$ & $.0586 \pm .0260$ & $.0653 \pm .0226$ & $4.0168 \pm 3.1656$   \\
            \midrule
            Neural Laplace & $.1063 \pm .0184$ & $.0426 \pm .0088$ & $.0250 \pm .0296$ & $.2764 \pm .1182$ \\
            Sig Neural Laplace &
            $\mathbf{.0540 \pm .0214}$ &
            $\mathbf{.0264 \pm .0035}$ &
            $\mathbf{.0076 \pm .0016}$ &
            $\mathbf{.2153 \pm .0787}$ \\
            \midrule
            Neural Flow ResNet &
            $.3360 \pm .1039$ &
            $.1675 \pm .0508$ &
            $\mathbf{.0513 \pm .0185}$ &
            $1.3076 \pm .2799$ \\
            Sig Neural Flow ResNet &
            $\mathbf{.2693 \pm .0343}$ &
            $\mathbf{.1343 \pm .0207}$ &
            $\mathbf{.0401 \pm .0235}$ &
            $\mathbf{.4501 \pm .1229}$ \\
            \bottomrule
        \end{tabular}
        }
    }
\end{table}

    \paragraph{Test set performance} Table \ref{tab:rmse-full-annotated} shows that our signature-based encoder consistently outperforms its RNN-based counterpart in terms of extrapolation performance, measured by the test RMSE on test sets. Moreover, we witness a global reduction of the standard deviation of the test RMSE, suggesting a more stable training w.r.t.\ the weight initialization. 

    \paragraph{Sensitivity analysis and ablation study} Table~\ref{tab:laplace-sensitivity} (and Table~\ref{tab:flowresnet-sensitivity} in Appendix~\ref{sec:appendix-add-num-exp}) demonstrates that increasing the depth substantially improves performance, highlighting the importance of the signature transform. Moreover, the learnt mapping $\Phi_{\theta}$ appears to be highly effective, supporting the intuition proposed by~\citet{bonnier2019deep} that such a mapping can mitigate issues arising when path information depends on higher-order signature terms.

    \begin{table}[!ht]
    \centering
    \caption{Sensitivity analysis and ablation study for Sig Neural Laplace. Test RMSE (mean $\pm$ std) over 5 runs. Best results bolded.}
    \label{tab:laplace-sensitivity}
    \renewcommand{\arraystretch}{1.3}
    \setlength{\tabcolsep}{7pt}
    {\small
    \begin{tabular}{cccccc}
    \toprule
    Study & Config & \shortstack{Delayed \\ Lotka-Volterra} & \shortstack{Spiral \\ DDE} & \shortstack{Delayed\\ Fitzhugh-Nagumo} & \shortstack{Delayed \\ Rössler} \\
    \midrule
    \multirow{3}{*}{\centering Depth $N$} 
    & $1$ & $.1592 \pm .0798$ & $.0316 \pm .0054$ & $.0148 \pm .0094$ & $.5130 \pm .2012$ \\
    & $2$ & $\mathbf{.0590 \pm .0146}$ & $.0348 \pm .0113$ & $.0178 \pm .0108$ & $.2394 \pm .1015$ \\
    & $3$ & $\mathbf{.0496 \pm .0243}$ & $\mathbf{.0261 \pm .0038}$ & $\mathbf{.0116 \pm .0048}$ & $\mathbf{.1682 \pm .0373}$ \\
    \midrule
    \multirow{2}{*}{\centering $\Phi_{\theta}$} 
    & $\times$ & $.1435 \pm .0889$ & $.0319 \pm .0081$ & $\mathbf{.0236 \pm .0115}$ & $.7351 \pm .1415$ \\
    & \checkmark & $\mathbf{.0605 \pm .0195}$ & $\mathbf{.0250 \pm .0015}$ & $\mathbf{.0177 \pm .0136}$ & $\mathbf{.2143 \pm .0840}$ \\
    \bottomrule
    \end{tabular}
    }
    \end{table}
 
    \paragraph{Training speed} Figure \ref{fig:training-loss-comparison_combined} (and Appendix \ref{sec:appendix-training-losses}) shows that the signature-based models achieve faster training, suggesting a more favorable loss landscape or informative gradient updates. It supports the claims made in the introduction that using the signature as an encoder may alleviate gradient issues encountered by RNNs \citep{pascanu2013difficulty}.

    \begin{figure}[!ht]
    \centering
        \floatconts
        {fig:training-loss-comparison_combined}
        {\caption{Training loss (log) vs. epochs on Spiral DDE averaged over 5 runs.}}
        {%
            \centering
                    \includegraphics[width=0.45\textwidth]{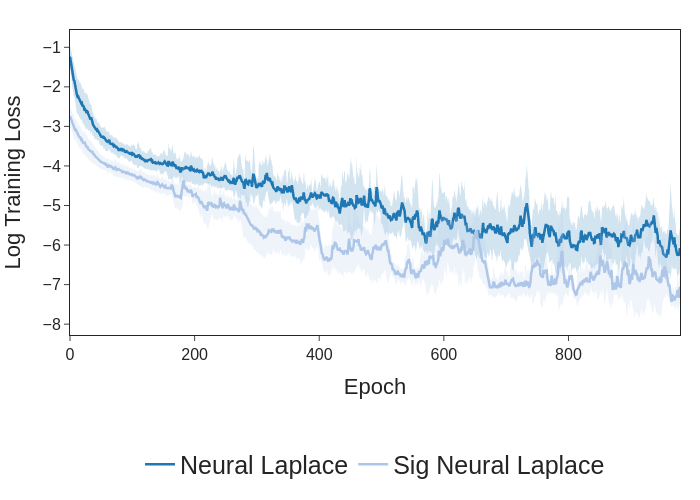}
                    \includegraphics[width=0.45\textwidth]{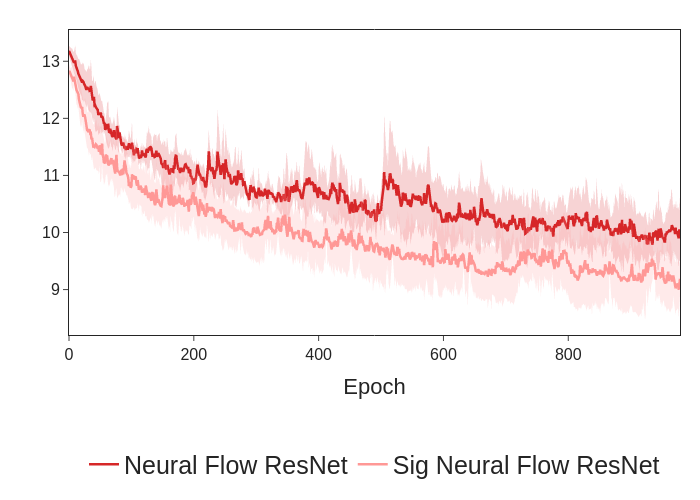}
                    
                    \vspace{-2.2em} 
                    
            }
\end{figure}

\begin{figure}
             \centering
            \includegraphics[width=0.38\textwidth]{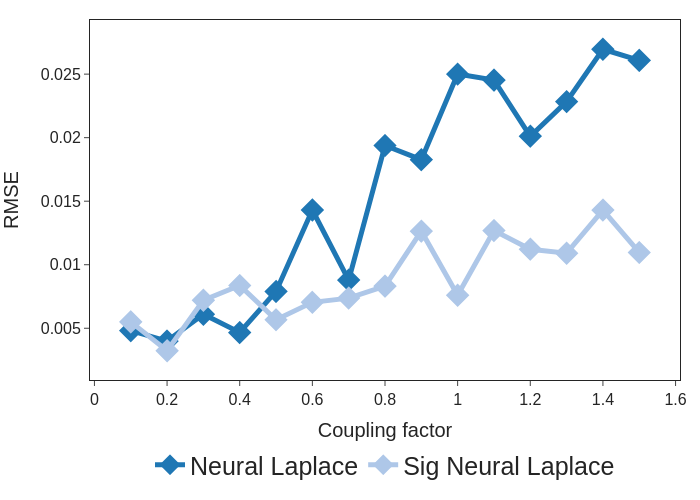}
            \caption{Test RMSE vs. coupling factor $\gamma$ averaged over 5 runs.}
            \label{fig:coupling}
    \end{figure}

    \paragraph{Multidimensional correlations}
    One reason signature-based encoders may perform better is their ability to capture multidimensional correlations within trajectories. To highlight this, we introduce a coupling factor $\gamma > 0$ in the Delayed Fitzhugh-Nagumo model and evaluate performance as $\gamma$ increases (see Appendix \ref{sec:appendix-coupling} for further details). Figure~\ref{fig:coupling} shows that signature-based model performance remains more stable than RNN-based, showcasing this advantage.

    \paragraph{Additional benchmarks} Appendix \ref{sec:appendix-add-num-exp} shows that using the signature is more computationally efficient. It further examines the models' performance when corrupting input trajectory with noise or sub sampling it.

    \section{Conclusion and future work}
    \label{sec:conclusion}

    In this paper, we proposed a unified approach that brings together diverse models for learning dynamical systems under a global encoder-decoder framework, and introduced a signature-based encoder for learning non-Markovian dynamics.
    We implemented this novel encoding scheme in two state-of-the-art models—Neural Laplace and Neural Flow ResNet —both capable of handling a broad class of differential equations, including delayed and stiff systems.
    To the best of our knowledge, previous work has predominantly relied on RNN-based encoders.

    We have presented a signature-based architecture that is fully compatible with the general encoder-decoder class of models introduced in this paper.
    We conducted comparative numerical experiments on synthetic benchmarks.
    Test datasets included a range of dynamical systems, covering multiple dimensions, delays and stiffness.
    Our results show that signature-based models yield significant improvements in both accuracy and  training stability, lead to faster and more computationally efficient training. To assess the impact of each encoder component, we performed an ablation study and tailored experiments to explain the observed performance gains. It is worth noting that other signature-based models, such as those introduced in the introduction \citep{kidger2020neuralcde, liao2021logsig, moreno2024rough}, could also have been employed as encoders. We leave this investigation for future work.
    
    The signature transform is drawing attention in the machine learning community, and we believe the performance improvement demonstrated in this study provides a valuable contribution to fields such as continuous time reinforcement learning \citep{yildiz2021rl}, control \citep{holt2023control} and time series forecasting. Furthermore, by showcasing the advantages of the signature transform over RNN-based encoders, this work highlights its potential to serve as a standard encoding mechanism for developing future models for learning dynamical systems. 


    \bibliography{bibliography}


    \appendix

    \section{Properties of the signature}
    \label{sec:appendix-sig-prop}

    In this section, we introduce the signature transform and its main properties on a broader class of paths. For convenience, we use the classical notation from the stochastic calculus literature.

    \begin{definition}[Path]
        \label{def:path}
        Let $a < b$ be two real numbers.  
        A (continuous) path is a (continuous) mapping $X = (X^1,\dots,X^d) : t \in [a,b] \rightarrow \mathbb{R}^d$.
    \end{definition}

    \begin{definition}[Total variation]
        \label{def:tv}
        The total variation of a continuous path $X : [a, b] \rightarrow \mathbb{R}^d$ is defined by $||X||_{\mathrm{TV}} \coloneqq \sup_{\mathcal{P} \subset [a, b]} \sum_{i=1}^{n-1} |X_{t_{i+1}} - X_{t_i}|$, where $\mathcal{P} \coloneqq (t_1, \dots, t_{n})$ denotes some partition of the interval $[a, b]$.
    \end{definition}
    We denote by $BV([a, b], \mathbb{R}^d)$ the set of continuous paths of bounded variation $X : [a, b] \rightarrow \mathbb{R}^d$ that satisfy $||X||_{\mathrm{TV}} < + \infty$.
    \begin{definition}[Signature transform]
        \label{def:sig-transform}
        Let $X \in BV([a, b], \mathbb{R}^d)$. The \textit{signature} of $X$ is then defined as the collection of iterated integrals
        \begin{align*}
        \mathrm{Sig}(X) &\coloneqq \left(
        \substack{
          \displaystyle \int \ldots \int \\
          \scriptstyle a < t_1 < \cdots < t_k < b
        }
        \ddroit X_{t_1} \otimes \cdots \otimes \ddroit X_{t_k}
        \right)_{k \geq 0} \\
        &= \left(\left(
        \substack{
        \displaystyle \int \ldots \int \\
        \scriptstyle a < t_1 < \cdots < t_k < b
        }
        \ddroit X^{i_1}_{t_1} \cdots \ddroit X^{i_k}_{t_k}
        \right)_{1 \leq i_1, \ldots, i_k \leq d} \right)_{k \geq 0},
        \end{align*}
        where the tensor product symbol $\otimes$ is a shorthand notation which represents the tensor nature of the signature's components, the $k = 0$ term is taken to be $1 \in \mathbb{R}$, and the integral is defined in the sense of Riemann–Stieltjes (or Young) integration \citep{friz2010multidimensional}.
        The \textit{truncated signature} of depth $N$ of $X$ is defined as
        \begin{align*}
        \mathrm{Sig}^N(X) \coloneqq \left(
        \substack{
        \displaystyle \int \ldots \int \\
        \scriptstyle a < t_1 < \cdots < t_k < b
        }
        \ddroit X_{t_1} \otimes \cdots \otimes \ddroit X_{t_k}
        \right)_{0 \leq k \leq N}.
        \end{align*}
    \end{definition}

    \begin{definition}[Tensor algebra]
        The tensor algebra of $\mathbb{R}^d$ is defined as
        \[
            T((\mathbb{R}^d)) \coloneqq \prod_{k=0}^{\infty} (\mathbb{R}^d)^{\otimes k}.
        \]
    \end{definition}
    The signature transform of a path $X$ lives in the tensor algebra space $T((\mathbb{R}^d))$, endowed with the tensor multiplication and the componentwise addition. It exhibits several properties, making it a good candidate for machine learning applications. First, with a time augmentation, the signature transform uniquely determines the encoded path.
    \begin{definition}[Time-augmented path]
        Let $X \in BV([a, b], \mathbb{R}^d)$. The time-augmented path $\tilde{X}$ is defined as $\tilde{X}_t \coloneqq (t,X_t)$ for all $t \in [a,b]$.
    \end{definition}
    \begin{proposition}[Uniqueness]
        \label{prop:uniqueness}If $\tilde{X}, \tilde{Y}$ denote the time-augmented paths, then $\mathrm{Sig}(\tilde{X}) = \mathrm{Sig}(\tilde{Y})$ implies that $\tilde{X} = \tilde{Y}$.
    \end{proposition}
    A proof of Proposition \ref{prop:uniqueness} can be found in \citet{hambly2010uniqueness}.
    In addition, the magnitude of the signature of order $N$ decays factorially.
    \begin{proposition}[Factorial decay]
        \label{prop:convergence-rate}Let $X \in BV([a, b], \mathbb{R}^d)$. \\
        Then for any $N \geq 0$,
        \[
            \left\lVert  \substack{
                \displaystyle \int \cdots \int \\
                \scriptstyle a < t_1 < \cdots < t_N < b
            }
                                     dX_{t_1} \otimes \cdots \otimes dX_{t_N}\right\rVert_{(\mathbb{R}^d)^{\otimes N}} \leq \frac{1}{N!}||X||_{\mathrm{TV}}^N
        \].
    \end{proposition}
    A proof of Proposition \ref{prop:convergence-rate} can be found in \citet{lyons2007differential}.
    \begin{remark}
        Using the definition of the signature transform, it is clear that $\mathrm{Sig}^N(X)$ lives in a space of dimension
        \[
            \sum_{i=0}^N d^i = \frac{d^{N+1}-1}{d-1},
        \]
        and therefore grows exponentially with the truncation order.
    \end{remark}
    The previous proposition is key to apply signatures to machine learning methods.
    Indeed, as the size of the $N^{th}$-truncated signature grows exponentially with $N$, it is necessary that $\mathrm{Sig}^N(X)$ is close to $\mathrm{Sig}(X)$ for a reasonable $N$ \citep{fermanian2021learning}.

    \begin{theorem}[Universal nonlinearity]
        \label{theorem:universal-nonlinearity} Let $K$ be a compact subset of $BV([a, b], \mathbb{R}^d)$.
        Let $f : K \rightarrow \mathbb{R}$ be a continuous function, and $\epsilon > 0$.
        Then there exists a linear operator $L : T((\mathbb{R}^d)) \rightarrow \mathbb{R}$ s.t.
        \[
            \forall \xbold \in K, \, |f(\tilde{X}) - L(\mathrm{Sig}(\tilde{X}))| < \epsilon
        \]
        with $\tilde{X}$ being the time-augmented path of $X$.

    \end{theorem}
    A proof of Theorem \ref{theorem:universal-nonlinearity} can be found in \citet{perez2018derivatives}.
    This final result is important for machine learning applications.
    However, two remarks should be made: 1) this is only an existence result and not a constructive one; 2) in practice, we work with truncated signatures, so that this result remains very theoretical.
    Indeed, linear models on the truncated signature may not be expressive enough, thus suggesting adding nonlinearities like the 1D convolution mentioned in Section \ref{sec:method}.
    \begin{proposition}[Time reparameterization invariance]
    \label{prop:time-reparam-invariance}
    Let $X \in BV([a, b], \mathbb{R}^d)$ and $\hat{X}_t := X_{\lambda(t)}$ with $\lambda : [a,b] \rightarrow [a,b]$ being a time reparameterization. Then $\mathrm{Sig}(\hat{X}) = \mathrm{Sig}(X)$.
    \end{proposition}
     A proof of Proposition \ref{prop:time-reparam-invariance} can be found in \citet{lyons1998differential}. Such proposition supports the fact that the signature transform is robust to irregular sampling, a claim already emphasized by \citet{moreno2024rough}.

    \section{Neural Laplace and Neural Flow ResNet}
    \label{sec:appendix-models-description}

    In this section, we adopt the notations used in the original papers.
    
    \paragraph{Neural Laplace \citep{holt2022neural}} The Neural Laplace model is made of three consecutive steps.
    First, a gated recurrent unit (GRU) \citep{cho-etal-2014-learning} $h_\gamma$ encodes the trajectory in a latent initial representation.
    Then the \textit{Laplace representation network} $g_\beta$ learns the dynamics in the Laplace domain, and finally maps it back to the temporal domain with an inverse Laplace transform (ILT).
    Precisely, given a stream of data $(\xbold_{t_1},\dots,\xbold_{t_n})$, the map $h_\gamma$ produces a latent initial condition representation vector $\mathbf{p} \in \mathbb{R}^l$, representing the encoded version of the input trajectory
    \begin{align*}
        \mathbf{p} = h_\gamma\left( (\xbold_{t_1}, t_1), \ldots, (\xbold_{t_n}, t_n) \right),
    \end{align*}
    which is then fed to the network $g_\beta$ to obtain the Laplace transform
    \begin{align*}
        \mathcal{L}\{\xbold\}(\mathbf{s}) = v\left( g_\beta(\mathbf{p}, u(\mathbf{s})) \right),
    \end{align*}
    where $\mathbf{s} \in \mathbb{C}^d$, $u$ is the stereographic projection and $v$ its inverse.
    
    Then, an inverse Laplace transform step is applied to both reconstruct and extrapolate the state estimate $\hat{\xbold}(t)$ at arbitrary time $t$.
    Within the framework introduced in Section \ref{sec:encoder-decoder}, the encoder is the network $h_\gamma$ that produces the vector $\mathbf{p}$, and the decoder is
    \begin{align*}
    (\mathbf{p}, t)
        \rightarrow \mathrm{ILT}\left(\mathcal{L}\{\xbold\}(\cdot),t\right)
    \end{align*}
    with $\mathcal{L}\{\xbold\}(\cdot) = v\left( g_\beta(\mathbf{p}, u(\mathbf{\cdot})) \right)$ and $\mathbf{p} = h_\gamma\left( (\xbold_{t_1}, t_1), \ldots, (\xbold_{t_n}, t_n) \right)$.

    \paragraph{Neural Flow ResNet \citep{bilos2021neural}} The Neural Flow ResNet architecture is more straightforward.
    Given a stream of data $(\xbold_{t_1},\dots,\xbold_{t_n})$, a GRU-flow encoder outputs an initial state $\mathbf{z_0} \in \mathbb{R}^l$.
    Then, the flow in the latent space is modelled by
    \begin{align*}
        F(t,\mathbf{z_0}) = \mathbf{z_0} + \varphi(t) g(t,\mathbf{z_0}),
    \end{align*}
    where $\varphi : \mathbb{R} \to \mathbb{R}^l$ (usually \texttt{tanh} function) and $g : \mathbb{R}^{l+1} \to \mathbb{R}^l$ is an arbitrary contractive neural network (\ie, $\text{Lip}(g) < 1$). Finally, a neural network is applied to map back to the original space.
    Within the framework introduced in Section \ref{sec:encoder-decoder}, the encoder is the aforementioned RNN-based encoder, and the decoder is the composition of the flow $F$ and the mapping back to the state space (original implementation).

    \section{Numerical implementation and hyperparameters}
    \label{sec:appendix-num-implementation}
    \begin{table}[htbp]
        \floatconts
        {tab:param_counts}
        {\caption{Number of parameters for the different methods and systems. Encoder refers to the encoder module (e.g., RNN-based or Signature transform); Total refers to the entire model.}}
        {%
            \begin{tabular}{lrrrr}
                \toprule
                \textbf{Method} & \multicolumn{2}{c}{\textbf{2D systems}} & \multicolumn{2}{c}{\textbf{3D systems}} \\
                \cmidrule(lr){2-3} \cmidrule(lr){4-5}
                & encoder & total & encoder & total \\
                \midrule
                NODE                   & 0         & 19450   & 0         & 19723  \\
                ANODE                  & 0         & 19723   & 0         & 19996   \\
                Neural Laplace         & 4391      & 21547   & 4454      & 25900   \\
                Sig Neural Laplace     & 3051      & 20207   & 4261      & 25707  \\
                Neural Flow ResNet     & 15064     & 18518   & 15272     & 18729   \\
                Sig Neural Flow ResNet & 8059      & 11513   & 12635     & 16092  \\
                \bottomrule
            \end{tabular}
        }
    \end{table}

    To ensure fair comparison, and as our purpose is to focus on the encoder choice, we tuned the models so that the number of parameters of the decoder remains constant across models (see Table \ref{tab:param_counts}).
    If it exists, we set the latent dimension to be 2.
    We use the Adam optimizer \citep{kingma2015adam} along with a learning rate of $10^{-3}$, a batch size of 128, and we train for 1000 epochs before taking the best model.
    Unless otherwise stated, we use the experimental code provided by \citet{holt2022neural} available on the official \href{https://github.com/samholt/NeuralLaplace}{Neural Laplace GitHub repository} to implement baseline models and do the experiments.
    Except for Neural Laplace model, the baselines are implemented according to the original papers.
    For the signature transform, we rely on the Pytorch-compatible library \verb|signatory| from \citet{bonnier2019deep} available on the official \href{https://github.com/patrick-kidger/Deep-Signature-Transforms}{Deep Signature Transforms Github repository}.
    We refer the reader to \citet{bonnier2019deep, reizenstein2020algorithm} for a tutorial on how to numerically compute the signature transform.
    All computations were performed on an \texttt{Intel(R) Core(TM) i7-14700K} CPU paired with an \texttt{NVIDIA RTX 4000} GPU, and \texttt{64 Go RAM}.

    \paragraph{Neural ODE \citep{chen2018neural}} We parameterize the ODE function $f$ using a 3-layer multilayer perceptron (MLP) with 136 hidden units and \texttt{tanh} activation functions. For training, the initial value is set to be the first trajectory value at the first observed time point. For extrapolation, it is the last observed trajectory value at the last observed time point.
    We use the `euler' solver with the semi-norm trick \citep{pmlr-v139-kidger21a}.

    \paragraph{Augmented Neural ODE \citep{dupont2019augmented}} We parameterize the ODE function $f$ as a 3-layer MLP with 136 hidden units and \texttt{tanh} activation functions.
    One dimension, initialized to zeros, is appended to the input. For training, the initial value is set to be the first trajectory value at the first observed time point. For extrapolation, it is the last observed trajectory value at the last observed time point.
    We again use the `euler' solver with the semi-norm trick \citep{pmlr-v139-kidger21a}.

    \paragraph{Neural Flow ResNet \citep{bilos2021neural}} We use the model of the original paper with 26 units.
    For the signature encoder, the augmentation $\phi_{\theta}$ is a 2-layer MLP with 25 hidden units. We set the number of features to be $4$, and enable both \verb|include_original| and \verb|include_time|. This results in an output size $e = 5 +$ (dimension of the system). The sliding window size (kernel size) is set to $m = 40$.
    The signature is truncated to depth 3, and the projection map $g_\xi$ is a linear layer.
    Both encoders output mean and standard deviation of the latent initial condition within the variational framework of the original paper, and the training loss for this model is the variational loss.

    \paragraph{Neural Laplace \citep{holt2022neural}} The Laplace representation model is a 3-layer MLP with 42 units per layer and \texttt{tanh} activations, as in the original paper.
    The original encoder is a GRU with 2 layers and 21 units, with a linear layer on the final hidden state.
    For the signature encoder, the augmentation $\phi_{\theta}$ is a 2-layer MLP with 25 hidden units. We set the number of features to be $4$, and enable both \verb|include_original| and \verb|include_time|. This results in an output size $e = 5 +$ (dimension of the system). The sliding window size (kernel size) is set to $m = 40$.
    The signature is truncated to depth 3, and the projection map $g_\xi$ is a linear layer.
    Both encoders output the initial representation $p$.

    \section{Sampling datasets and evaluation method}
    \label{sec:appendix-sampling-de}



    For the experiments, we use the exact same method as \citet{holt2022neural}.
    We evaluate model performance in an extrapolation setting by splitting each sampled trajectory into two equal parts \([0, T/2]\) and \([T/2, T]\).
    The first half is used for encoding, and the second half for prediction.
    For each dynamical system, 1,000 trajectories are sampled, each initialized from a different initial condition.
    We use a train-validation-test split of 80:10:10.
    For repeated experiments with the same method, we set a different random seed.
    Finally, all datasets are normalized using statistics computed from the training set only, to prevent data leakage (between training and test sets).
    To sample our datasets, we use the \texttt{ddeint} numerical solver \citep{zulko_ddeint}.

    \paragraph{Delayed Lotka-Volterra} We simulate trajectories of the delayed Lotka–Volterra system using a fixed delay of \( \tau = 0.1 \).
    The trajectories are computed over the interval \( t \in [2, 30] \) with 1,000 time points, then subsampled uniformly to 200 time points.
    We vary the initial conditions over a grid in \([0.1, 2] \times [0.1, 2] \).

    \paragraph{Spiral DDE} We simulate trajectories of the spiral delay differential equation system using a fixed delay of \( \tau = 2.5 \).
    The matrix $A$ is set to $
    \begin{bmatrix}
        -1 & 1  \\
        -1 & -1
    \end{bmatrix}$.
    The trajectories are computed over the interval \( t \in (0, 20] \) with 1,000 time points, then subsampled uniformly to 200 time points.
    We vary the initial conditions over a grid in \([-2, 2] \times [-2, 2] \).

    \paragraph{Delayed Fitzhugh-Nagumo} We simulate trajectories of the delayed Fitzhugh–Nagumo system using a fixed delay of \( \tau = 1 \).
    The system parameters are set to \( a = 0.5 \), \( b = 0.8 \), \( \varepsilon = 0.02 \), \( I = 0.5 \).
    The trajectories are computed over the interval \( t \in [2, 30] \) with 1,000 time points, and uniformly subsampled to 200 time points.
    We vary the initial conditions over a grid in \([-5, 5] \times [-5, 5] \).

    \paragraph{Delayed Rössler} We simulate trajectories of the delayed Rössler system using a fixed delay of \( \tau = 2.5 \).
    The system parameters are set to \( a = 0.2 \), \( b = 0.2 \), and \( c = 4.5 \).
    The trajectories are computed over the interval \( t \in [2, 20] \) with 1,000 time points, and uniformly subsampled to 200 time points.
    We vary the initial conditions over a grid in \([0.1, 1.5]^3\).

    \section{Coupling factor experiment}
    \label{sec:appendix-coupling}
    
    For the numerical experiment presented in Section \ref{fig:coupling}, we introduce a coupling factor $\gamma > 0$ in the Delayed Fitzhugh-Nagumo model
    \[
    \begin{aligned}
    \dot{x_1}(t) &= x_1(t) - \frac{x_1^3(t)}{3} - \gamma x_2(t - \tau) + I, \\
    \dot{x_2}(t) &= \varepsilon \left(\gamma x_1(t) + a - b x_2(t) \right)
    \end{aligned}
    \]
    As the coupling parameter $\gamma$ increases, component-wise correlations play a more significant role, since $\gamma$ directly governs the strength of the coupling. We therefore expect the performance of the signature-based model to degrade less than that of the original model.
    To test this hypothesis, we evaluate the extrapolation performance of both models across varying values of $\gamma$. The experimental setup is identical to that described in Appendices \ref{sec:appendix-num-implementation} and \ref{sec:appendix-sampling-de}, with the only difference being that $\gamma$ is allowed to vary. 
    
    \section{Training losses}
    \label{sec:appendix-training-losses}


    \begin{figure}[htbp]
        \floatconts
        {fig:training-loss-comparison-lotka-volterra}
        {\caption{Training loss (log) vs. epochs averaged over 5 runs for Delayed Lotka Volterra system. Left : Neural Laplace models, right : Neural Flow ResNet models.}}
        {%
            \centering
            \resizebox{\textwidth}{!}{%
                \begin{minipage}{\textwidth}
                    \includegraphics[width=0.49\textwidth]{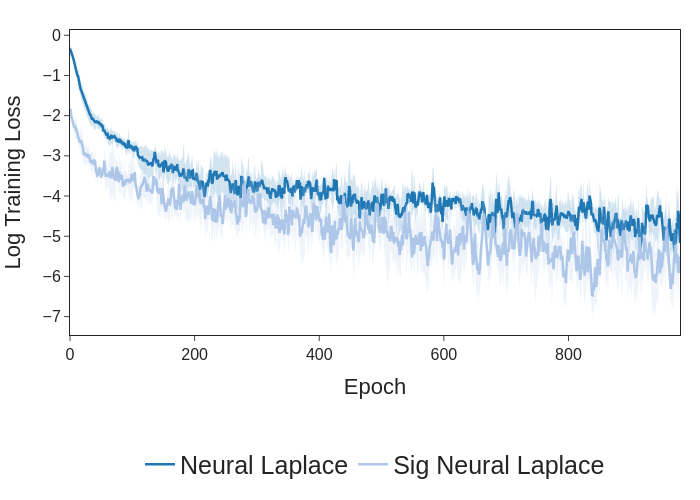}
                    \hfill
                    \includegraphics[width=0.49\textwidth]{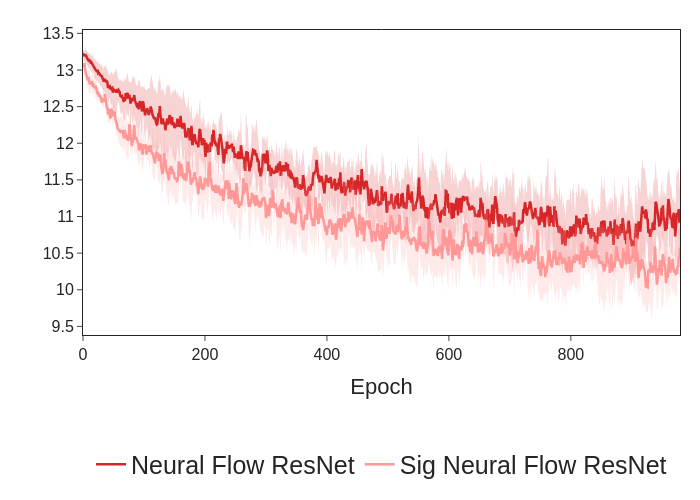}
                \end{minipage}
            }
        }
    \end{figure}

    \begin{figure}[htbp]
        \floatconts
        {fig:training-loss-comparison_spiral-dde}
        {\caption{Training loss (log) vs. epochs averaged over 5 runs for Spiral DDE system. Left : Neural Laplace models, right : Neural Flow ResNet models.}}
        {%
            \centering
            \resizebox{\textwidth}{!}{%
                \begin{minipage}{\textwidth}
                    \includegraphics[width=0.49\textwidth]{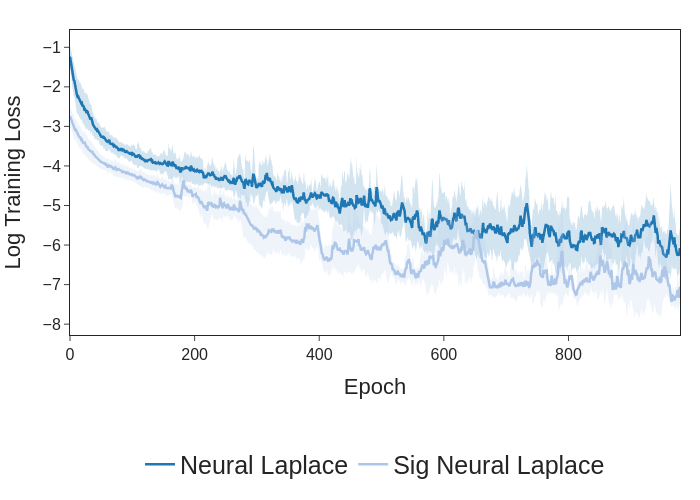}
                    \hfill
                    \includegraphics[width=0.49\textwidth]{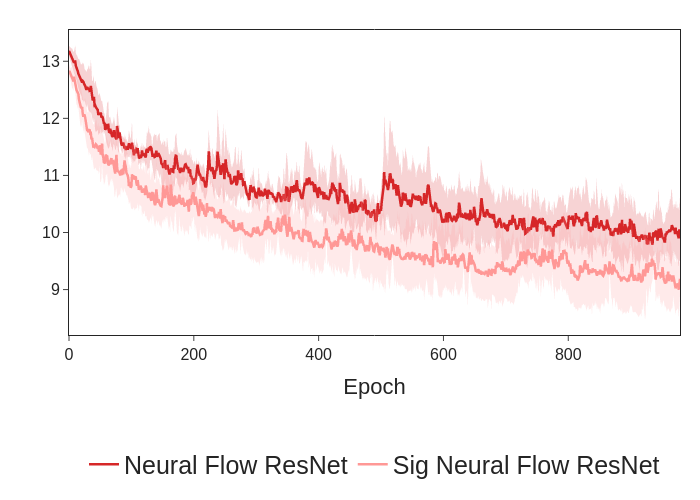}
                \end{minipage}
            }
        }
    \end{figure}

    \begin{figure}[htbp]
        \floatconts
        {fig:training-loss-comparison_fitzhugh-nagumo}
        {\caption{Training loss (log) vs. epochs averaged over 5 runs for Delayed Fitzhugh-Nagumo system. Left : Neural Laplace models, right : Neural Flow ResNet models.}}
        {%
            \centering
            \resizebox{\textwidth}{!}{%
                \begin{minipage}{\textwidth}
                    \includegraphics[width=0.49\textwidth]{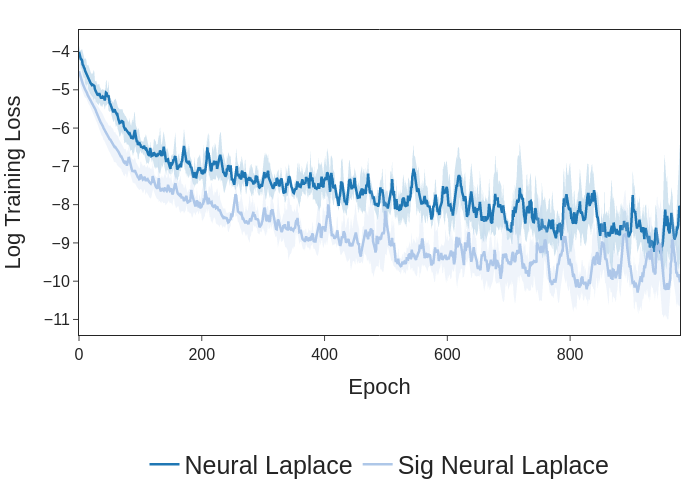}
                    \hfill
                    \includegraphics[width=0.49\textwidth]{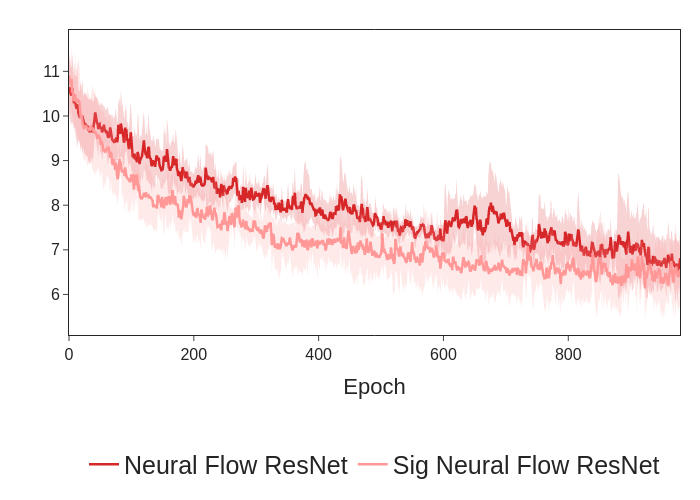}
                \end{minipage}
            }
        }
    \end{figure}

    \begin{figure}[htbp]
        \floatconts
        {fig:training-loss-comparison_rossler}
        {\caption{Training loss (log) vs. epochs averaged over 5 runs for Delayed Rössler system. Left : Neural Laplace models, right : Neural Flow ResNet models.}}
        {%
            \centering
            \resizebox{\textwidth}{!}{%
                \begin{minipage}{\textwidth}
                    \includegraphics[width=0.49\textwidth]{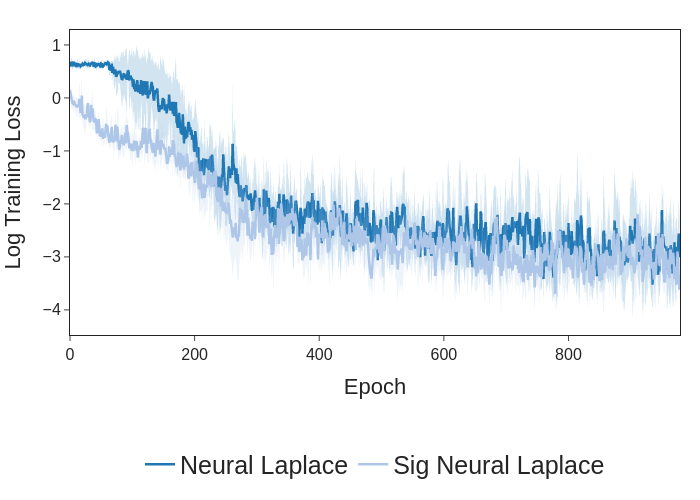}
                    \hfill
                    \includegraphics[width=0.49\textwidth]{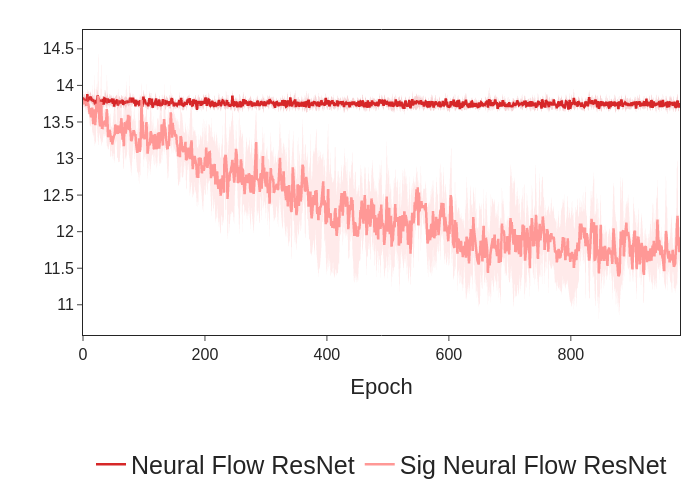}
                \end{minipage}
            }
        }
    \end{figure}

    \newpage

    \section{Additional benchmarks}
    \label{sec:appendix-add-num-exp}

    \begin{table}[htbp]
    \centering
    \caption{Sensitivity analysis and ablation study for Sig Neural Flow ResNet. Test RMSE (mean $\pm$ std) over 5 runs. Best results bolded.}
    \label{tab:flowresnet-sensitivity}
    \renewcommand{\arraystretch}{1.3}
    \setlength{\tabcolsep}{7pt}
    {\small
    \begin{tabular}{cccccc}
    \toprule
    Study & Config. & \shortstack{Delayed \\ Lotka-Volterra} & \shortstack{Spiral \\ DDE} & \shortstack{Delayed \\ Fitzhugh-Nagumo} & \shortstack{Delayed \\ Rössler} \\
    \midrule
    \multirow{3}{*}{\centering\arraybackslash \shortstack{Depth $N$}} 
    & $1$ & $.3423 \pm .1028$ & $.1985 \pm .0404$ & $.0601 \pm .0200$ & $.6897 \pm .1276$ \\
    & $2$ & $\mathbf{.2409 \pm .0665}$ & $.2026 \pm .0572$ & $.0403 \pm .0210$ & $.5810 \pm .0924$ \\
    & $3$ & $\mathbf{.2685 \pm .0717}$ & $\mathbf{.1892 \pm .0531}$ & $\mathbf{.0313 \pm .0081}$ & $\mathbf{.4939 \pm .0856}$ \\
    \midrule
    \multirow{2}{*}{\centering\arraybackslash \shortstack{$\Phi_{\theta}$}} 
    & $\times$ & $\mathbf{.2486 \pm .0354}$ & $.2283 \pm .0629$ & $.0504 \pm .0148$ & $.8470 \pm .2476$ \\
    & \checkmark & $\mathbf{.2833 \pm .0770}$ & $\mathbf{.1664 \pm .0331}$ & $\mathbf{.0393 \pm .0100}$ & $\mathbf{.5073 \pm .2330}$ \\
    \bottomrule
    \end{tabular}
    }
    \end{table}

    \begin{table}[htbp]
        \floatconts
        {tab:epoch-duration}
        {\caption{Mean epoch duration (in seconds). Best results per group bolded.}}
        {%
                {\small
            \renewcommand{\arraystretch}{1.2} 
            \setlength{\tabcolsep}{4pt} 
            \begin{tabular}{lcccc}
                \toprule
                \textbf{Method} &
                \shortstack{Delayed \\Lotka-Volterra} &
                \shortstack{Spiral \\DDE} &
                \shortstack{Delayed \\Fitzhugh-Nagumo} &
                \shortstack{Delayed \\Rössler}  \\
                \midrule
                ANODE (euler)   & $.4683$  & $.4720$ & $.4754$ & $.5388$ \\
                NODE (euler) & $.4693$ & $.4734$ & $.4907$ & $.5369$ \\
                \midrule
                Neural Laplace & $ .0907$ & $.0927$ & $.0907$ & $.1200$ \\
                Sig Neural Laplace &
                $\mathbf{.0756}$ &
                $\mathbf{.0773}$ &
                $\mathbf{.0751}$ &
                $\mathbf{.1093}$ \\
                \bottomrule
            \end{tabular}
            }
        }
    \end{table}

    We measured the mean epoch duration with a batch size of 128, and we observe that Sig Neural Laplace is roughly 15\% faster than Neural Laplace for 2D systems (Table \ref{tab:epoch-duration}).

    \begin{table}[htbp]
        \floatconts
          {tab:lotka-noise}
          {\caption{Test RMSE on Delayed Lotka-Volterra across increasing noise levels $\varepsilon$, averaged over 5 runs.}}
          {%
            {\small
            \renewcommand{\arraystretch}{1.2}
            \setlength{\tabcolsep}{6pt}
            \begin{tabular}{lcccc}
            \toprule
            \textbf{Method} & $\varepsilon = 0$ & $\varepsilon = 0.02$ & $\varepsilon = 0.05$ & $\varepsilon = 0.1$ \\
            \midrule
            ANODE & $.4472 \pm .0641$ & $.3214 \pm .0391$ & $.4930 \pm .0465$ & $.5120 \pm .0282$ \\
            NODE & $.6061 \pm .2521$ & $.4700 \pm .1963$ & $.6422 \pm .1847$ & $.6236 \pm .1551$ \\
            \midrule
            Neural Laplace & $.1063 \pm .0184$ & $.0955 \pm .0174$ & $\mathbf{.1118 \pm .0102}$ & $\mathbf{.1632 \pm .0140}$ \\
            Sig Neural Laplace & $\mathbf{.0541 \pm .0214}$ & $\mathbf{.0514 \pm .0169}$ & $\mathbf{.1159 \pm .0010}$ & $.1813 \pm .0135$ \\
            \midrule
            Neural Flow ResNet & $.3360 \pm .1038$ & $\mathbf{.2774 \pm .0609}$ & $.3208 \pm .0670$ & $.3823 \pm .0487$ \\
            Sig Neural Flow ResNet & $\mathbf{.2693 \pm .0343}$ & $\mathbf{.2546 \pm .0393}$ & $\mathbf{.2728 \pm .0313}$ & $\mathbf{.3108 \pm .0298}$ \\
            \bottomrule
            \end{tabular}
            }
          }
    \end{table}

    \begin{table}[htbp]
    \floatconts
      {tab:spiral-noise}
      {\caption{Test RMSE on Spiral DDE across increasing noise levels $\varepsilon$, averaged over 5 runs.}}
      {%
        {\small
        \renewcommand{\arraystretch}{1.2}
        \setlength{\tabcolsep}{6pt}
        \begin{tabular}{lcccc}
        \toprule
        \textbf{Method} & $\varepsilon = 0$ & $\varepsilon = 0.02$ & $\varepsilon = 0.05$ & $\varepsilon = 0.1$ \\
        \midrule
        ANODE & $.0447 \pm .0097$ & $.0631 \pm .0082$ & $.1108 \pm .0042$ & $.2113 \pm .0038$ \\
        NODE & $.0590 \pm .0255$ & $.0724 \pm .0223$ & $.1182 \pm .0170$ & $.2162 \pm .0076$ \\
        \midrule
        Neural Laplace & $.0430 \pm .0107$ & $.0508 \pm .0120$ & $.0791 \pm .0089$ & $\mathbf{.1394 \pm .0122}$ \\
        Sig Neural Laplace & $\mathbf{.0265 \pm .0044}$ & $\mathbf{.0380 \pm .0038}$ & $\mathbf{.0749 \pm .0034}$ & $\mathbf{.1404 \pm .0068}$ \\
        \midrule
        Neural Flow ResNet & $.1902 \pm .0191$ & $.1860 \pm .0059$ & $.2406 \pm .0203$ & $\mathbf{.2244 \pm .0283}$ \\
        Sig Neural Flow ResNet & $\mathbf{.1317 \pm .0149}$ & $\mathbf{.1322 \pm .0184}$ & $\mathbf{.1572 \pm .0292}$ & $\mathbf{.1955 \pm .0380}$ \\
        \bottomrule
        \end{tabular}
        }
      }
    \end{table}

    \begin{table}[htbp]
        \floatconts
        {tab:fitzhugh-noise}
        {\caption{Test RMSE on Delayed Fitzhugh--Nagumo across increasing noise levels $\varepsilon$, averaged over 5 runs. Best results per group bolded.}}
        {%
                {\small
            \renewcommand{\arraystretch}{1.2}
            \setlength{\tabcolsep}{6pt}
            \begin{tabular}{lcccc}
                \toprule
                Model                  & $\varepsilon=0$            & $\varepsilon=0.02$         & $\varepsilon=0.05$         & $\varepsilon=0.1$          \\
                \midrule
                ANODE                  & $.1001 \pm .0531$          & $.1015 \pm .0412$          & $.1314 \pm .0330$          & $.1649 \pm .0318$          \\
                NODE                   & $.0653 \pm .0226$          & $.0628 \pm .0195$          & $.0949 \pm .0254$          & $.1412 \pm .0069$          \\
                \midrule
                Neural Laplace         & $.0250 \pm .0296$          & $.0343 \pm .0240$          & $.0650 \pm .0194$          & $\mathbf{.1071 \pm .0095}$ \\
                Sig Neural Laplace     & $\mathbf{.0076 \pm .0016}$ & $\mathbf{.0224 \pm .0007}$ & $\mathbf{.0554 \pm .0017}$ & $\mathbf{.1070 \pm .0024}$ \\
                \midrule
                Neural Flow ResNet     & $\mathbf{.0513 \pm .0184}$ & $\mathbf{.0500 \pm .0135}$ & $\mathbf{.0742 \pm .0063}$ & $\mathbf{.1175 \pm .0099}$ \\
                Sig Neural Flow ResNet & $\mathbf{.0401 \pm .0235}$ & $\mathbf{.0451 \pm .0173}$ & $\mathbf{.0777 \pm .0100}$ & $.1426 \pm .0085$ \\
                \bottomrule
            \end{tabular}
            }
        }
    \end{table}

    \begin{table}[htbp]
        \floatconts
        {tab:fitzhugh-seq-length}
        {\caption{Test RMSE on Fitzhugh--Nagumo across decreasing input sequence length $n$, averaged over 5 runs. Best results per group bolded.}}
        {%
                {\small
            \renewcommand{\arraystretch}{1.2}
            \setlength{\tabcolsep}{6pt}
            \begin{tabular}{lcccc}
                \toprule
                Model                  & $n=100$                    & $n=80$                     & $n=70$                     & $n=50$                     \\
                \midrule
                ANODE                  & $.1001 \pm .0531$          & $.1070 \pm .0523$          & $.0954 \pm .0530$          & $.0975 \pm .0559$          \\
                NODE                   & $.0653 \pm .0226$          & $.0826 \pm .0338$          & $.0528 \pm .0172$          & $.0557 \pm .0187$          \\
                \midrule
                Neural Laplace         & $.0250 \pm .0296$          & $.0528 \pm .0196$          & $.0537 \pm .0230$          & $.1120 \pm .0229$          \\
                Sig Neural Laplace     & $\mathbf{.0076 \pm .0016}$ & $\mathbf{.0159 \pm .0036}$ & $\mathbf{.0197 \pm .0025}$ & $\mathbf{.0449 \pm .0106}$ \\
                \midrule
                Neural Flow ResNet     & $.0513 \pm .0185$          & $.0484 \pm .0121$          & $\mathbf{.0453 \pm .0143}$ & $\mathbf{.0687 \pm .0244}$ \\
                Sig Neural Flow ResNet & $\mathbf{.0401 \pm .0235}$ & $\mathbf{.0388 \pm .0225}$ & $\mathbf{.0446 \pm .0212}$ & $\mathbf{.0684 \pm .0216}$ \\
                \bottomrule
            \end{tabular}
            }
        }
    \end{table}

    \section{Dataset plots}
    \label{sec:appendix-de-plots}

    This section presents visualisations of predictions over the interval $[T/2, T]$, generated by Neural Laplace and Neural Flow ResNet models on trajectories randomly selected from the test set, both with and without the signature-based encoder. For reference, the full ground-truth trajectory is also included. (A)NODE models are omitted to highlight the effect of the signature-based encoder on qualitative performance. In addition, for the Delayed Rössler system, we exclude Neural Flow ResNet trajectories, as their divergence makes the plots unreadable.


    \begin{figure}[htbp]
        \includegraphics[width=\linewidth]{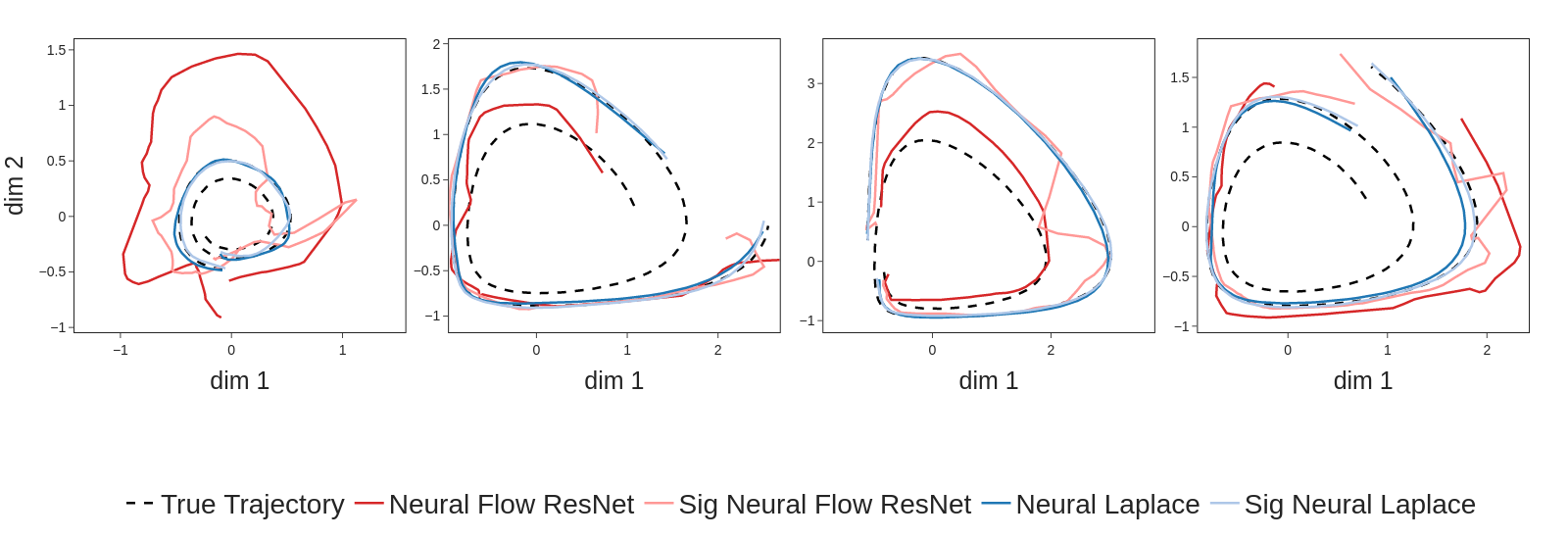}
        \caption{Delayed Lotka-Volterra randomly sampled test trajectory.}
    \end{figure}

    \begin{figure}[htbp]
        \includegraphics[width=\linewidth]{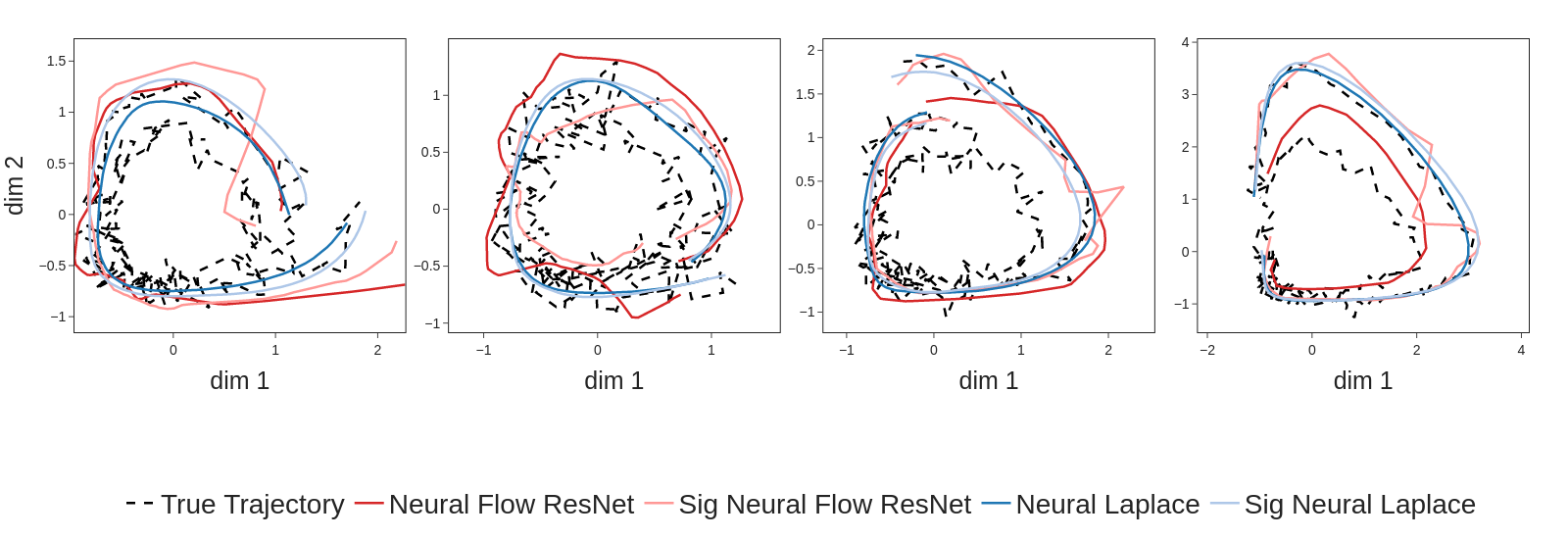}
        \caption{Delayed Lotka-Volterra randomly sampled test trajectory, with gaussian noise $\mathcal{N}(0,0.1)$ added.}
    \end{figure}

    \begin{figure}[htbp]
        \includegraphics[width=\linewidth]{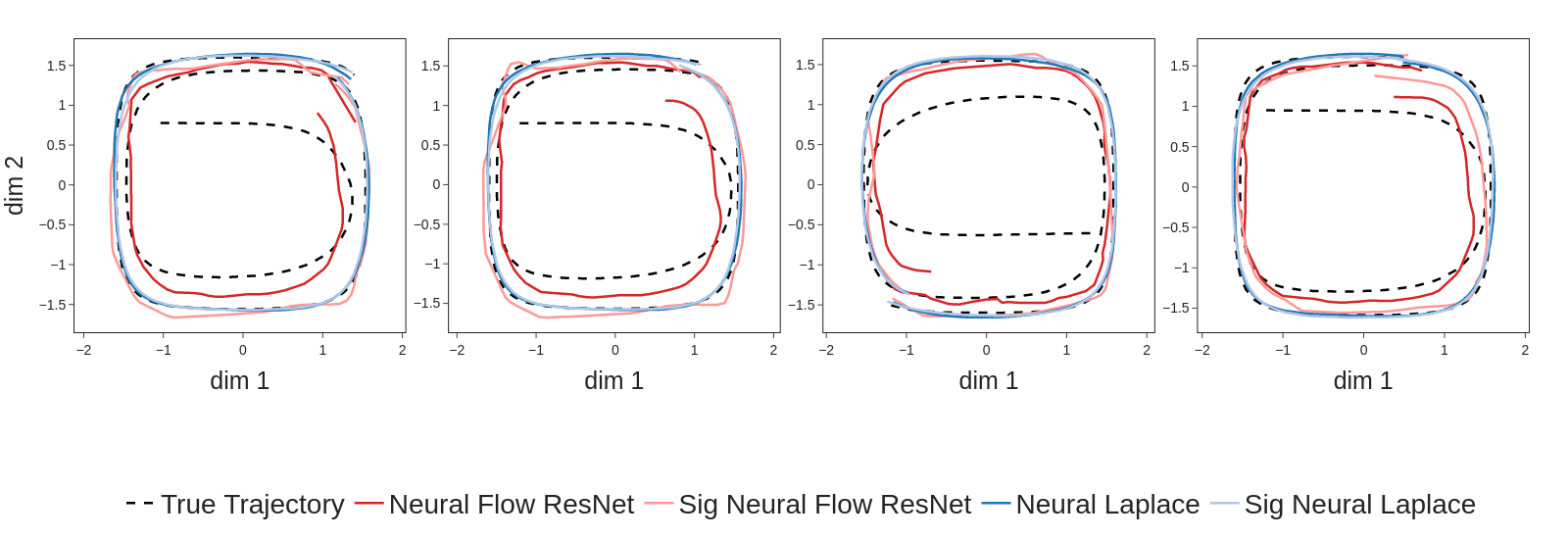}
        \caption{Spiral DDE randomly sampled test trajectory.}
    \end{figure}

    \begin{figure}[htbp]
        \includegraphics[width=\linewidth]{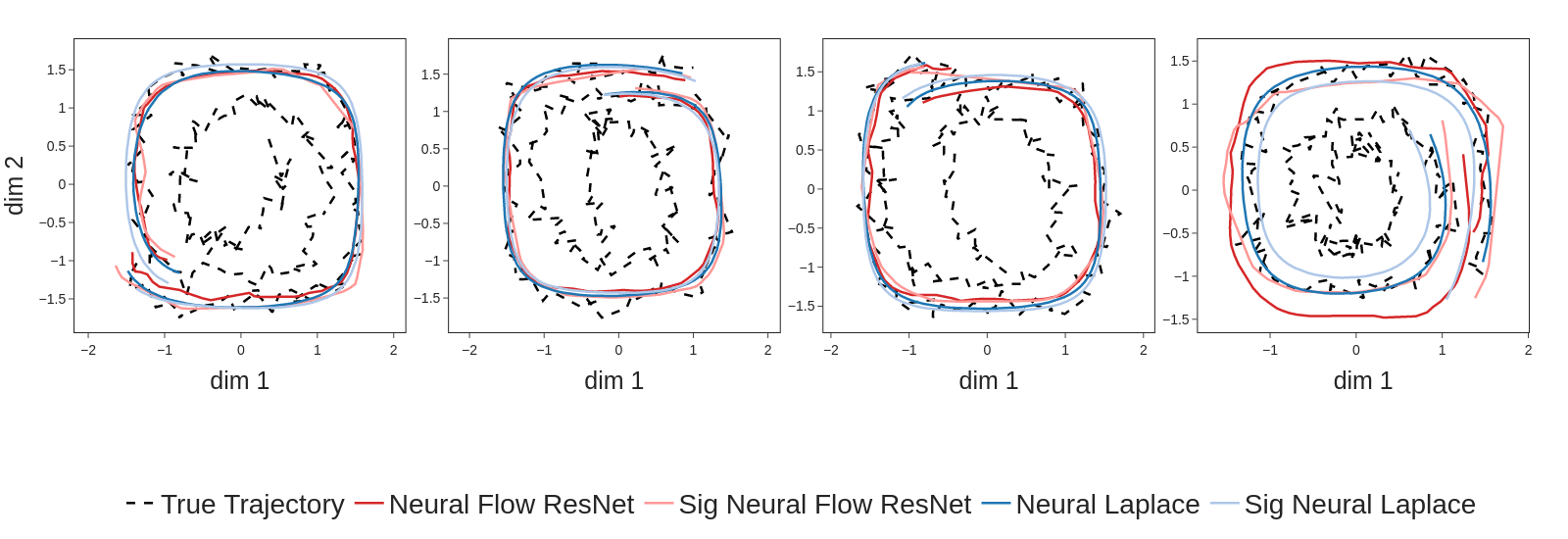}
        \caption{Spiral DDE randomly sampled test trajectory, with gaussian noise $\mathcal{N}(0,0.1)$ added.}
    \end{figure}

    \begin{figure}[htbp]
        \includegraphics[width=\linewidth]{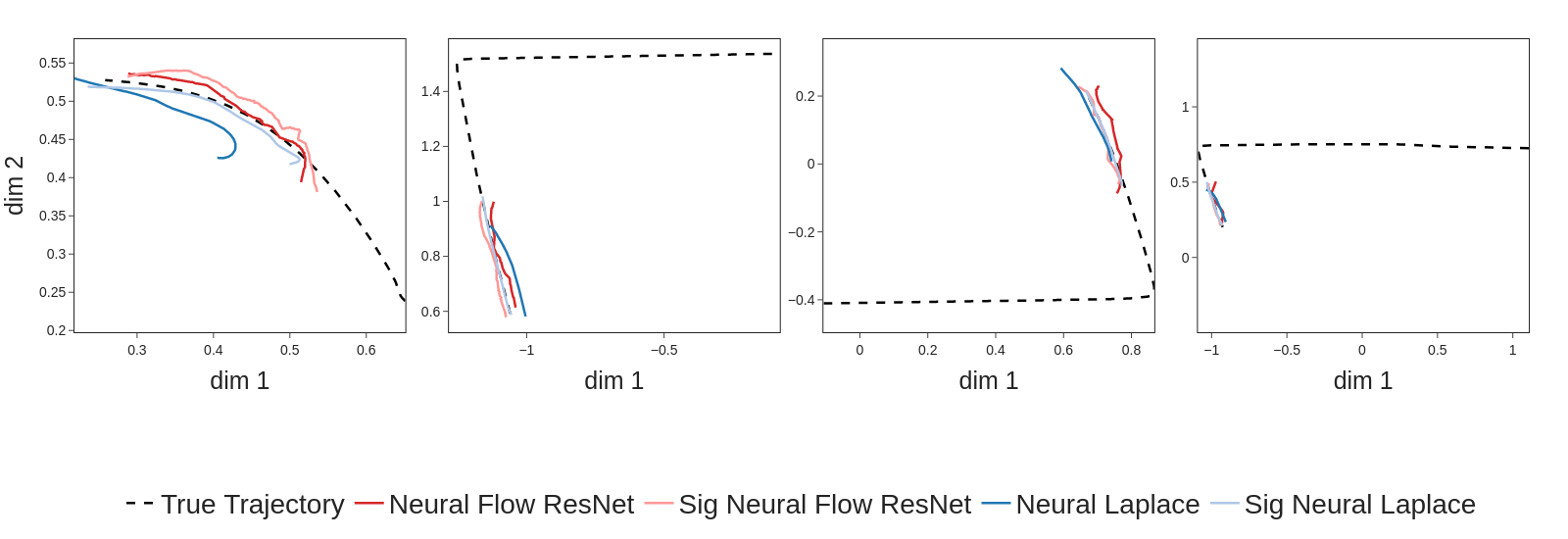}
        \caption{Delayed Fitzhugh-Nagumo randomly sampled test trajectory.}
    \end{figure}

    \begin{figure}[htbp]
        \includegraphics[width=\linewidth]{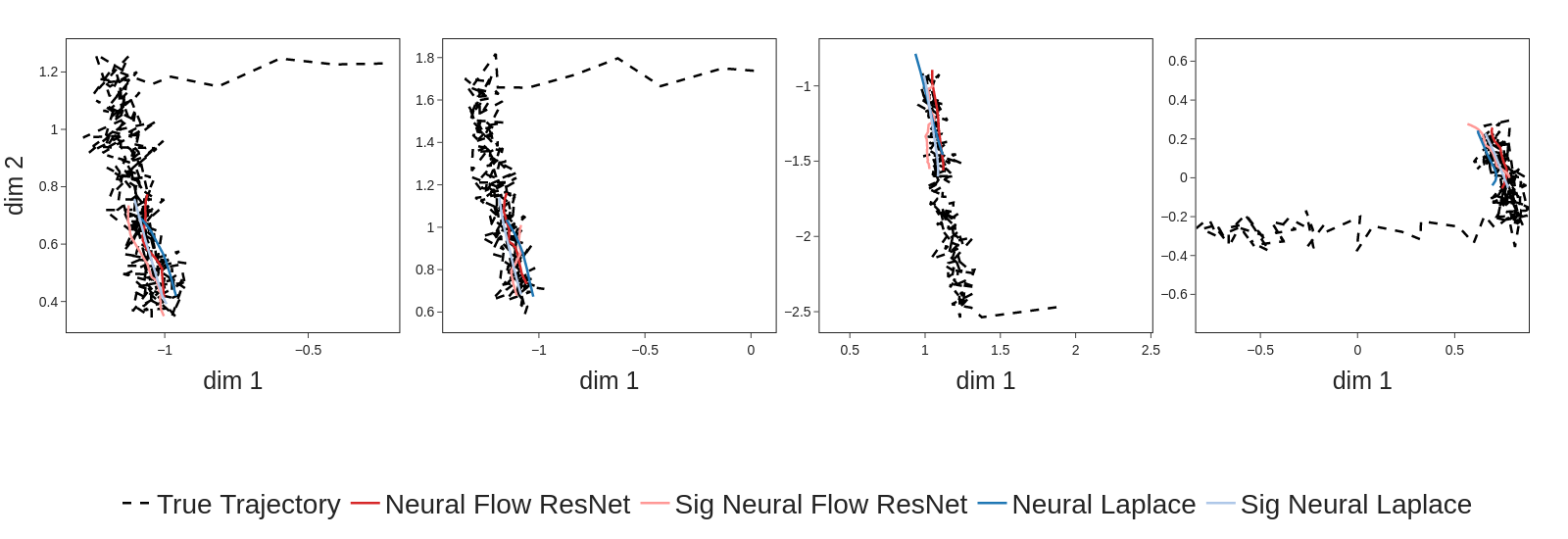}
        \caption{Delayed Fitzhugh-Nagumo randomly sampled test trajectory, with gaussian noise $\mathcal{N}(0,0.05)$ added.}
    \end{figure}

    \begin{figure}[htbp]
        \includegraphics[width=\linewidth]{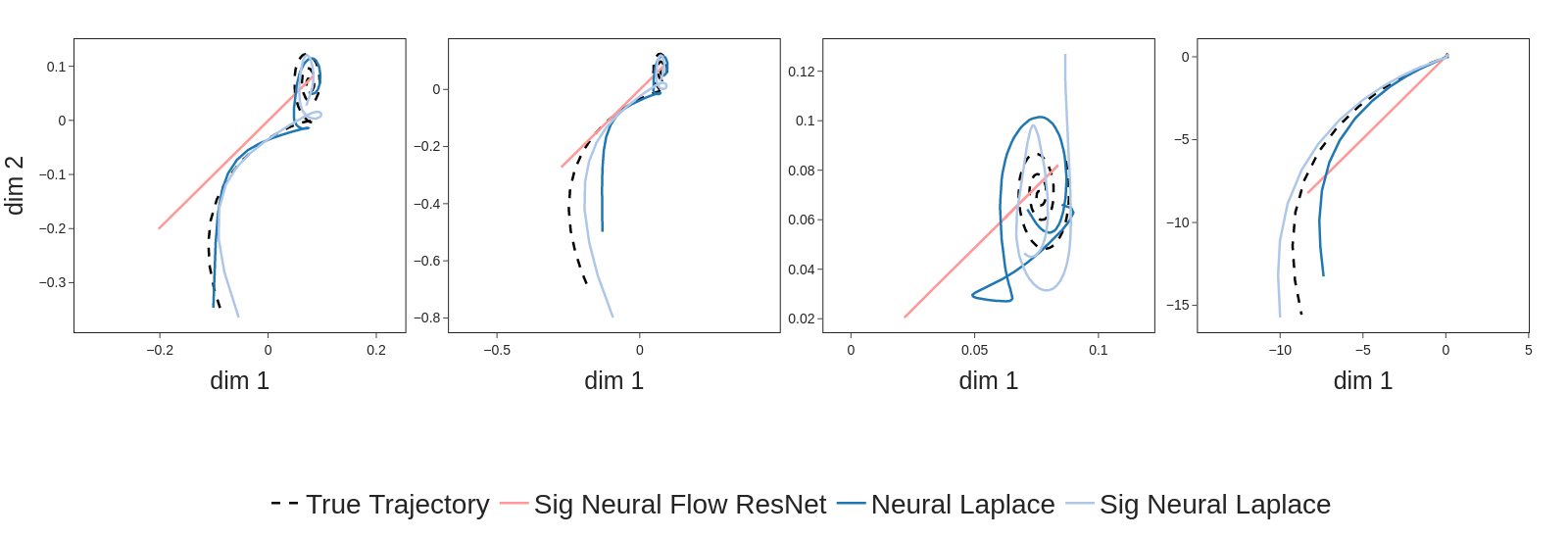}
        \caption{Delayed Rössler (first two dimensions) randomly sampled test trajectory.}
    \end{figure}

    \begin{figure}[htbp]
        \includegraphics[width=\linewidth]{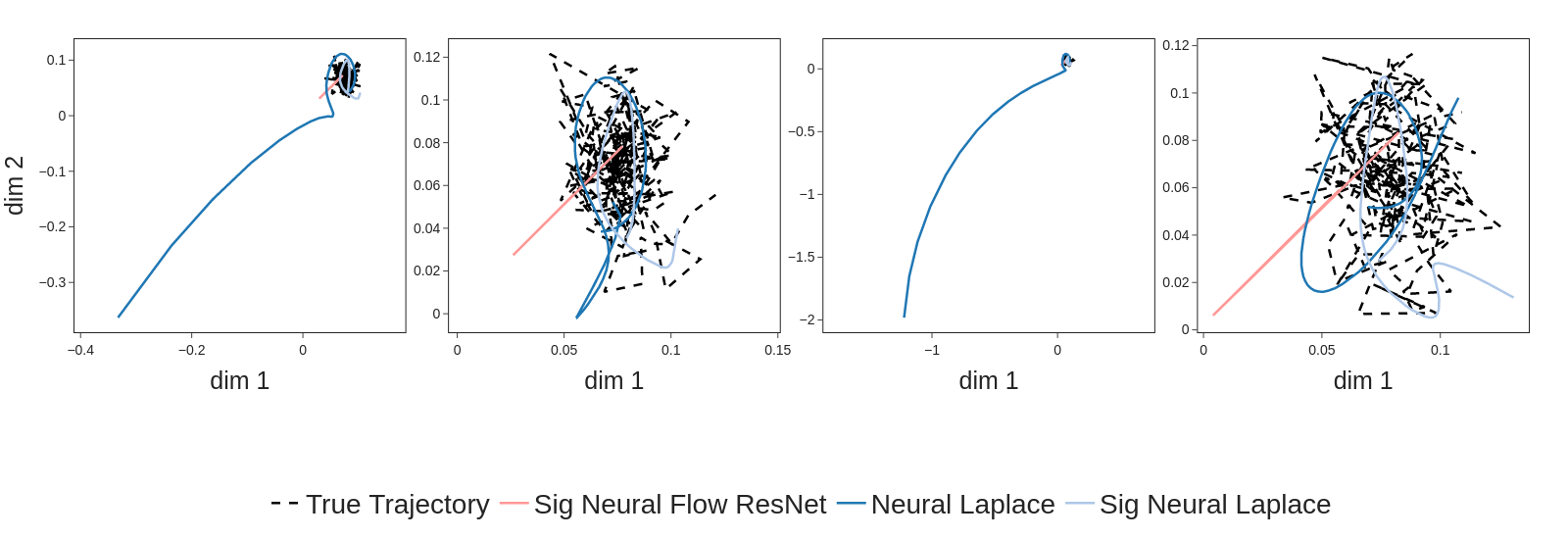}
        \caption{Delayed Rössler (first two dimensions) randomly sampled test trajectory, with gaussian noise $\mathcal{N}(0,0.01)$ added.}
    \end{figure}
\end{document}